\documentclass[letterpaper]{article} 
\usepackage{aaai25}  
\usepackage{times}  
\usepackage{helvet}  
\usepackage{courier}  
\usepackage[hyphens]{url}  
\usepackage{graphicx} 
\usepackage{natbib}  
\usepackage{caption} 
\usepackage{algorithm}
\usepackage{algorithmic}

\usepackage{newfloat}
\usepackage{listings}
\DeclareCaptionStyle{ruled}{labelfont=normalfont,labelsep=colon,strut=off} 
\floatstyle{ruled}
\newfloat{listing}{tb}{lst}{}
\floatname{listing}{Listing}
\usepackage{bm}
\usepackage{amsmath,amssymb}
\usepackage{booktabs} 
\usepackage{xcolor}
\usepackage{multirow}

\title{Enhancing Noise-Robust Losses for Large-Scale Noisy Data Learning}
\author {
    Max Staats \textsuperscript{\rm 1,2} \thanks{Staats is the corresponding author.},
    Matthias Thamm \textsuperscript{\rm 2},
    Bernd Rosenow \textsuperscript{\rm 2}
}
\affiliations {
    \textsuperscript{\rm 1}ScaDS.AI Dresden/Leipzig, Germany\\
   \textsuperscript{\rm 2} Institut f\"{u}r Theoretische Physik, Universit\"{a}t
  Leipzig,  Br\"{u}derstrasse 16, 04103 Leipzig, Germany \\
    \{staats,thamm\}@itp.uni-leipzig.de, rosenow@physik.uni-leipzig.de
}

\begin{document}

\maketitle

\begin{abstract}
Large annotated datasets inevitably contain noisy labels, which poses a major challenge for training deep neural networks as they easily memorize the labels.
Noise-robust loss functions have emerged as a notable strategy to counteract this issue, but it remains challenging to create a robust loss function which is not susceptible to underfitting.  
Through a quantitative approach, this paper explores the limited overlap between the network output at initialization and regions of non-vanishing gradients of bounded loss functions in the initial learning phase.
Using these insights, we address underfitting of several noise robust losses with a novel method denoted as \emph{logit bias},  
which adds a real number $\epsilon$ to the logit at the position of the correct class. 
The \emph{logit bias} enables these losses to achieve state-of-the-art results, even on datasets like WebVision, consisting of over a million images from 1000 classes. 
In addition, we demonstrate that our method can be used to determine 
optimal parameters for several loss functions  
-- without having to train networks.
Remarkably, our method determines
the hyperparameters based on the number of classes, resulting in loss functions which require zero dataset or noise-dependent parameters.
\end{abstract}

%

\section{Introduction}

Supervised deep learning depends on high-quality labeled data for effective pattern recognition and accurate predictions  \cite{goodfellow2016deep}. 
In real-world datasets, however, there is often label noise - erroneous or unclear labels due to human error or incomplete annotations  \cite{liang2022review}. Such noise can drastically impair the effectiveness of deep learning models, which often operate under the assumption of pristine labels \cite{song2022learning}.  Therefore, it is important to develop robust deep-learning algorithms that can efficiently learn from noisy datasets.

One effective approach to navigate label noise lies in employing noise-robust loss functions. These loss functions, notable for their model-agnostic nature, seamlessly integrate with any deep learning paradigm.
The existing literature highlights their ability to improve the robustness and generalization ability of deep learning models under noisy conditions
\cite{ghosh2017robust,zhang2018generalized,wang2019symmetric,amid2019robust,ma2020normalized,zhou2021asymmetric,englesson2021generalized}.

A majority of these loss functions are bounded to prevent the learning of mislabeled examples. From a theoretical point of view, bounded losses have a higher robustness to noise if they belong to the class of symmetric losses \cite{ghosh2017robust}. Nonetheless, it has been suggested that such symmetry could be overly constraining \cite{zhou2021asymmetric}, with functions like the Mean Absolute Error (MAE)  leaning towards underfitting.
Reflecting this, many contemporary loss functions do not satisfy this symmetry condition \cite{zhou2021asymmetric,englesson2021generalized}.

In this paper, we quantitatively explore how the vanishing derivatives of bounded loss functions impact their learning behavior. 
According to our findings, the cause of underfitting is the limited overlap between the output values of an initialized network and the region where the derivative of a particular bounded loss function is nonzero.
To counteract this, we add a real number, $\epsilon$, to the logit corresponding to the correct class label. This subtle adjustment restores the overlap between network outputs and the region of sufficiently large derivatives of the loss, enabling e.g. the MAE loss to surpass the Cross Entropy loss on datasets like Cifar-100, even in the absence of label noise.
Impressively, this approach requires only a single constant $\epsilon$, which is determined by the number of classes, providing an effectively parameter-free method. Other loss functions like the generalized cross entropy (genCE)  \cite{zhang2018generalized} and NF-MAE (a combination of normalized focal loss (NF) with MAE) \cite{ma2020normalized} are also able to learn the WebVision dataset with the help of the logit bias.

Furthermore, our description of the early learning phase enables us to calculate suitable hyperparameters for other loss functions like genCE and NCE-AGCE (a combination of Normalized Cross Entropy and Asymmetric Generalized Cross Entropy) \cite{zhou2021asymmetric}, allowing them to show their potential for an arbitrary number of classes  
without the need of tuning their parameters first, e.g., by an expensive grid search. The method we propose 
is intended as a first step towards a universal framework that is capable of noise robust learning across varied numbers of classes
without requiring hyperparameter fine-tuning. The need for such a method is underscored by our observation that none of the proposed loss functions that are noise resistant on the Cifar-10 dataset are capable of learning the WebVision dataset.

In summary, our paper 
(i) quantitatively describes how the initial learning phase of a newly initialized network is contingent upon the dataset's class count; (ii) explores the limitations of bounded losses in multi-class datasets and introduces the \emph{logit bias} technique, enabling MAE to consistently deliver competitive or even superior results across benchmarks like Fashion-MNIST, Cifar-10, Cifar-100, and WebVision -- without hyperparameters; (iii) enables other noise-robust loss functions to learn the WebVision dataset using either the logit bias or by determining their hyperparameters without training a single network.

All code for reproducing the data and creating the figures in this paper is open source and available under \cite{code}.

\section{Related Work} 

Label noise in training data is a pervasive challenge that has attracted much attention in recent years \cite{liang2022review,song2022learning}. One strategy for addressing it is data cleaning, aiming to filter out mislabeled samples from the training dataset.
 To identify noisy instances, \cite{xiao2015learning} employs a probabilistic model to capture the relationship between images, labels, and noise. Other approaches utilize an auxiliary neural network, trained on curated data, to clean the main dataset \cite{veit2017learning,lee2018cleannet}. Yet, an overzealous curation can sometimes be counterproductive, as eliminating too many samples might degrade model performance \cite{khetan2017learning}, compared to retaining some corrupted instances. 

Another approach is estimating the noise transition matrix, which depicts the likelihood of mislabeling across classes. This matrix can be incorporated directly into the loss function \cite{han2018masking} or inferred throughout training \cite{goldberger2017training,sukhbaatar2014learning}, mitigating the consequences of mislabeling. A variation of this strategy involves dynamically adjusting the weights of samples during training. For example, \cite{reed2014training} adjust labels based on the network's current predictions, and \cite{ren2018learning} evaluate label trustworthiness based on the gradient induced by a given example.
Furthermore, \cite{thulasidasan2019combating} prompt the network to predict the likelihood of the example being correct, enabling fine-tuned loss adjustments.

Another avenue entails constraining the minimum training loss, emphasizing that an optimal scenario that avoids learning from incorrect labels will incur a finite loss \cite{toner2023label}. 
 Self-supervised methods that iteratively adjust labels by extracting information directly from data structures — via dynamic label learning \cite{chen2020simple} and contrastive learning \cite{hendrycks2019using,zheltonozhskii2022,ghosh2021contrastive,xue2022investigating,yi2022learning} — have also been shown to improve model generalization in noisy conditions.

Opting for a noise-robust loss function can complement and enhance many of the strategies described above.
In \cite{ghosh2017robust} it has been shown that bounded and symmetric losses are noise-tolerant.
The belief that symmetric losses are prone to underfitting
\cite{zhang2018generalized,zhou2021asymmetric} led to the development of alternative loss functions. \cite{feng2021can} augment the Cross Entropy loss to render it bounded, and \cite{lyu2019curriculum} introduce the \emph{curriculum loss}, acting as a tight upper bound to the 0-1 loss. Meanwhile, \cite{xu2019l_dmi} advocate for a non-bounded loss based on information-theoretic arguments.

Various combinations have emerged to take advantage of the strengths of multiple losses.
The generalized Cross Entropy (genCE) \cite{zhang2018generalized} synergizes   MAE with the Cross Entropy loss (CE) using a Box-Cox transformation, in an attempt to combine the fast initial learning performance of CE with the noise robustness of MAE.  On the other hand, symmetric Cross Entropy (symCE) uses a  linear combination of these two losses \cite{wang2019symmetric}.
Active-Passive Losses employ linear combinations of MAE  with a normalized version of either CE or the Focal loss \cite{ma2020normalized}. Lastly, the Bi-Tempered Logistic loss (biTemp) \emph{tempers}  the exponential and logarithmic functions 
to create robust loss function similar to the cross entropy \cite{amid2019robust}.

\section{Theoretical Considerations}\label{Sec:AnCons} 

Consider a classification task with $K$ classes, an example space defined as $X$, and associate labels $\{1,2,...K\}$. We define a loss function, $\mathcal{L}$,  as symmetric if 
\begin{equation}
    \sum_{y=1}^K \mathcal{L}(f(x),y) = C \quad \forall x \in X \quad \forall  f\colon \mathbb{R}^n\mapsto \mathbb{R}^K,
\end{equation}
where  $C\in\mathbb{R}$ is a fixed constant. Symmetric loss functions have been demonstrated to exhibit robustness against uniform and class-conditional noise \cite{ghosh2017robust}.  The MAE, for instance, serves as a notable example of a symmetric loss.
Such losses are often bounded, meaning that constant $C\in\mathbb{R}$ exists such that $\mathcal{L}(\bm{a} ,\bm{y}) < C$, regardless of label $\bm{y}$ and network output $\bm{a}$. 

A heuristic understanding of this theory can be gained by recognizing that wrongly labeled examples tend to deviate significantly from their class in  feature space.
In a scenario where these outliers can accumulate unbounded loss, the gradient  adjusts the decision boundaries to accommodate even the most extreme outliers.
In contrast, a bounded loss ensures that the gradient decreases at greater distances from the decision boundary, preventing overfitting these outliers. 

The downside of this approach is a potential slowdown in learning speed, as legitimate examples could be misinterpreted as outliers, with no gradient available to adjust the network weights for accurate classification.  
The symmetric MAE loss embodies this scenario — it is robust to noise \cite{ghosh2017robust}, yet underperforms on datasets like Cifar-100 \cite{zhang2018generalized}.

This section focuses on the early learning stage of a generic neural network and attempts to determine why certain noise-robust loss functions underfit on multiclass datasets 
\cite{zhang2018generalized}. In addition, we want to identify strategies to address this problem. To explore the underlying causes, we imagine a freshly initialized network with random weights $\mathbf{W}$ drawn from a zero-mean distribution.
We focus on the logit distribution for each neuron $z_i$ in the final layer, denoted as $p_z(z_i)$.  

Owing to inherent symmetry, we obtain $p_z(z_i)=p_z(z_j)\equiv p_z(z)$. 
The central limit theorem suggests that when the width $N$ of the last hidden layer is infinite, the logit distribution $p_z$ corresponds to a normal distribution with zero mean \cite{neal1994priors,lee2017deep}. As $N$ is typically quite large, $p_z(z)$ can indeed be well approximated by a zero-mean normal distribution.

 Under conventional training regimes, where the weights are initialized with variances given by the inverse of the layer output size, initial neuron pre-activations have approximately a unit variance\footnote{It is worth noting that there have been instances where smaller initializations have been recommended \cite{yang2021tuning}.}. Therefore, we adopt a $\mathcal{N}(0,1)$ distribution for the logits $z_i$. Though our results primarily consider this distribution, they can be easily generalized for other variances. For an empirical demonstration of the distribution under standard initializations, we refer to the appendix.

\begin{figure}[t]
  \centering
  \includegraphics[width=0.8\linewidth]{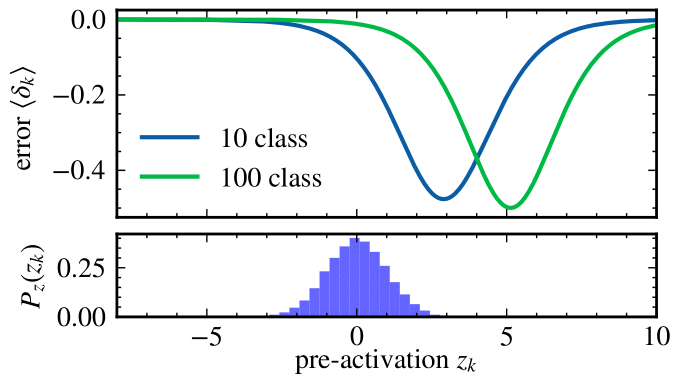}
  \caption{Analysis of the average error for the neuron corresponding to the correct label, determined for the MAE loss. Increasing the number of classes from 10 to 100 leads to a notable shift in the range where the average error $\langle \partial_{z_k} \mathcal{L} \rangle$ is non-vanishing. However, the logit distribution of a newly initialized network (shown in the histogram at the bottom) remains class-count invariant. This mismatch leads to diminished gradients, stalling the learning due to tiny errors. }
    \label{fig:mae_error}
  \end{figure}

When training neural networks using backpropagation, the error $\delta$ associated with neuron $n$ can be expressed as
\begin{equation}
    \delta_n = \frac{\partial \mathcal{L}}{\partial {z_n} }   = 
    \sum_{j=1}^K \frac{\partial \mathcal{L}}{\partial {a_j}} a_j ( \delta[j-n] -a_n),
\end{equation} 
where $K$ denotes the number of classes and $\delta[j-n]$ the Kronecker delta, $\delta[0]=1$ and $\delta[m]=0$ for $m\neq0$. Moreover, we assume that the network output $a_i$ at position $i$ is defined as $a_i=\exp(z_i)/ \sum_j \exp(z_j)$ where $z_i$ is the logit at output position $i$ meaning that we assume the softmax activation for the final layer.

This error plays a central role in backpropagation. Specifically, it determines the magnitude of the gradient descent updates, expressed as $\nabla_{W_l} \mathcal{L} = \bm{\delta} \cdot \bm{a}_{l-1}^T$. An error of zero results in a zero gradient, causing learning to stall.

From our previous discussions, we established that the distribution of $z_i$ corresponds to a normal distribution,
 $\mathcal{N}(0,1)$. Consequently, in the initial learning phases the average error of neuron $n$ as a function of the logit $z_n$ of that neuron is 
 given by:
\begin{multline}\label{eq:back_prop_er} 
  \langle \delta_n \rangle(z_n) = \Bigg\langle \sum_{j=1}^K \frac{\partial \mathcal{L}}{\partial {a_j}} \frac{\exp(z_j)}{\sum_i \exp(z_i)} \times \\
  \Bigg( \delta[j-n]   -  \frac{\exp(z_n)}{\sum_i \exp(z_i)} \Bigg) 
  \Bigg\rangle_{z_{i}, i \neq n }  .
\end{multline}

To gain insights, we visualize the average error, $\langle \delta_k \rangle$, in the neuron $k$ corresponding to the correct label as a function of its pre-activation. This involves computing the average with respect to $z_{i \neq k}$ and assigning a static value to $z_k$. This perspective offers a snapshot of typical error values in a newly initialized network.
Using the MAE as an example, Fig.~\ref{fig:mae_error} shows that with 10 classes, the range over which the average error $\langle \delta_k \rangle$  notably deviates from zero overlaps significantly with the logit distribution of a network at initialization. However, this overlap diminishes with 100 classes as the error depends on the number of classes. 

This illustration underscores the challenge: as the number of classes increases, gradients vanish under the MAE loss.  This phenomenon has implications for all bounded loss functions, which assign small gradients for outliers.  Importantly, what is deemed an "outlier" varies with the class count. We will later illustrate how this makes learning on the WebVision dataset infeasible for certain noise-robust loss functions optimized for up to 100 classes. 

Relevant to this discussion, \cite{ma2020normalized} categorize the active component of a loss function as the contribution that optimizes the output in the correct neuron. This component has been identified as essential to successful fitting.  
The optimization within the correct neuron turns out to be the linchpin of the fitting process, especially since the errors in the incorrect neurons decrease with increasing number of classes:
 already for $100$ classes, the initial activation of an ``incorrect'' neuron in the final layer is very close to the optimal value of zero, with an average value of $0.01$, leading to rather small errors in these neurons. This accentuates the need to focus primarily on the error in the ``correct'' neuron, $\delta_k$.

\begin{table*}[t]
\centering
\begin{tabular}{l l c l c}
    \multicolumn{1}{l}{loss} & \multicolumn{1}{l}{definition} & \multicolumn{1}{c}{ parameters} & \multicolumn{1}{l}{output error $\delta_n$} & \multicolumn{1}{c}{Ref.}\\
    \toprule 
    CE      & $ - \log( a_k )$ 
            & 0 & $a_n - \delta[n-k]$ & -\\ \addlinespace[0.2em]
    MAE     & $2(1- a_k)$ 
            & 0 & $2 a_k (  a_n - \delta[n-k])$ & -\\ \addlinespace[0.3em]
    genCE   & $q^{-1}(1- ( a_k )^q)$ 
            & 1 & $(a_k)^{q}  ( a_n - \delta[n-k] )$ &  \cite{zhang2018generalized}\\ \addlinespace[0.3em]
    biTemp   & no analytic form
            & 0 & no analytic form &        
            \cite{amid2019robust} \\
    NCE   & $\frac{ \log(a_k) }{ \sum_i \log(  a_i ) }$
            & 0 & see appendix  &        
            \cite{ma2020normalized} \\ \addlinespace[0.3em]
    NF   & $\frac{ \log \left( (1- a_k)^{0.5} a_k  \right) }{ \sum_i \log \left(  (1- a_i)^{0.5} a_i \right) }$
            & 0 & see appendix  &        
            \cite{ma2020normalized} \\ \addlinespace[0.3em]
    AGCE   & $[(b+1)^q-(b +a_k)^q]/q$  
            & 2 & $a_k (a_n -\delta[n-k] )(b+a_n)^{q-1}$  &        
            \cite{zhou2021asymmetric}\\ \addlinespace[0.3em]
    NF-MAE   & $\alpha \; {\rm NF}  + \beta \; {\rm MAE} $
            & 1 & see appendix  &        
            \cite{ma2020normalized} \\ \addlinespace[0.3em]
    NCE-MAE   & $\alpha \; {\rm NCE}  + \beta \; {\rm MAE} $
            & 1 & see appendix  &        
            \cite{ma2020normalized} \\ \addlinespace[0.3em]

    NCE-AGCE   & $\alpha  \; {\rm NCE} + \beta \; {\rm AGCE} $
            & 4 & see appendix &        
            \cite{zhou2021asymmetric}\\ \addlinespace[0.3em]
    \bottomrule
    \end{tabular}
\caption{Overview of the examined loss functions, with the neuron associated with the correct label indexed by $k$. The reported number of parameters excludes values that we keep constant for all datasets. Instead, it focuses on values actively adjusted based on the dataset under consideration. Parameter values utilized in our simulations, sourced from respective publications, are detailed in the appendix.}
\label{Tab:LossFncts}
\end{table*}

\begin{figure}[t] %
  \centering
  \includegraphics[width=\linewidth]{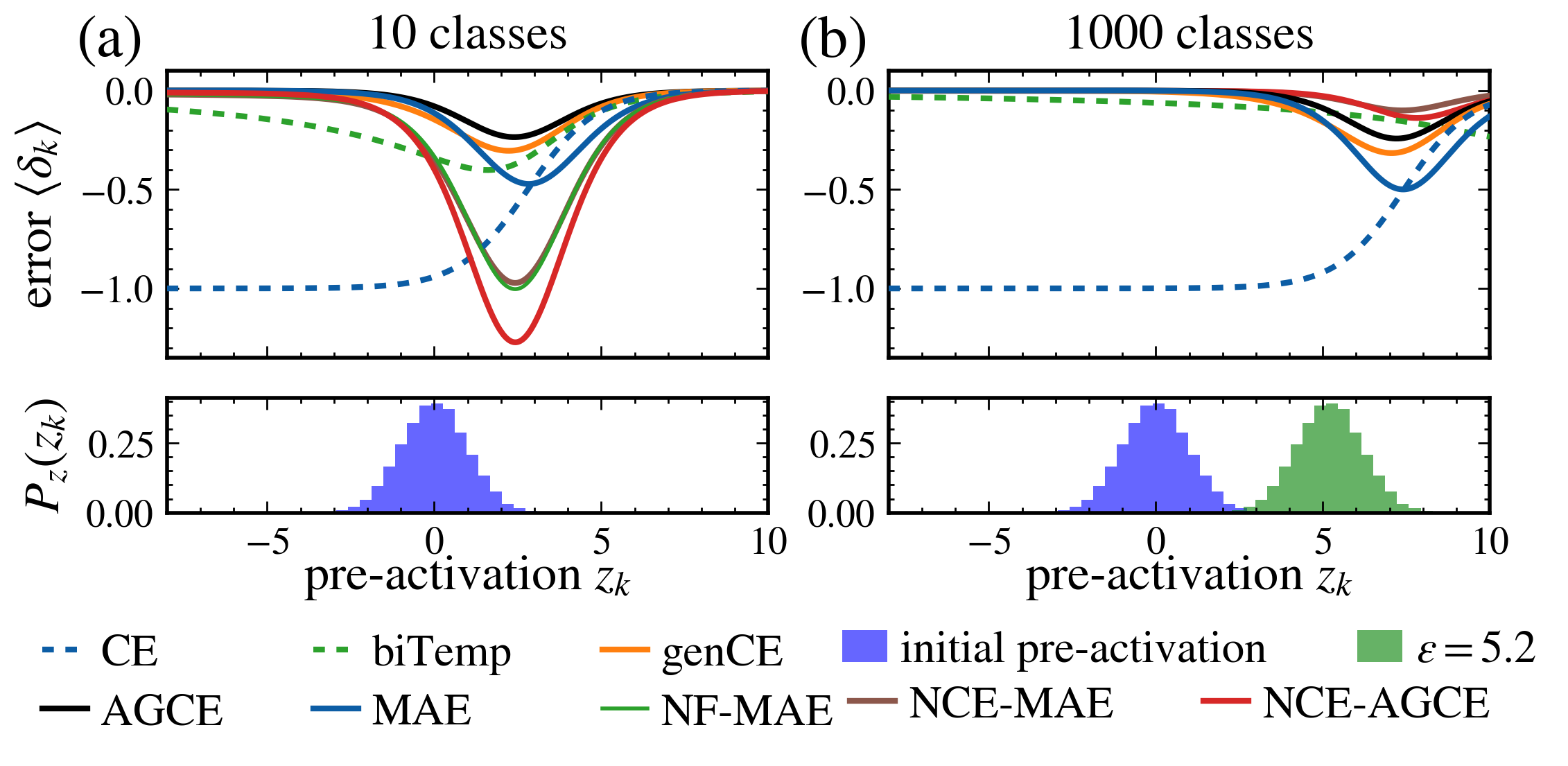}
  \caption{The average error $\langle \delta_k \rangle$ of the final layer's correct neuron $k$, a determinant factor for the magnitude of a gradient descent update,  plotted against  the pre-activation $z_k$,  for a network at initialization. Panel (a) depicts a ten-class learning scenario, showcasing a pronounced overlap between the region where learning is possible  (characterized by large negative $\langle \delta_k \rangle$) and the logit distribution from a randomly initialized network (blue histogram). In contrast, in a scenario with 1000 classes shown in panel (b), this overlap is not present for the bounded loss functions. Rather than employing a tailed loss --- exemplified by biTemp (delineated by the dashed green line) --- our approach involves shifting the $z_k$ distribution into the learning-possible range by adding a bias to the correct neuron's pre-activation (green histogram).  We note that the average error of NCE-AGCE is rescaled by a factor of $0.5$ in panel a) to enhance visual clarity.  }
    \label{fig:learning_curves}
  \end{figure}

\subsection{Comparison of various loss functions}

To address the underfitting described above for datasets with a moderate number of classes, prior work has proposed two main strategies: (i) Implementing an unbounded loss of the form 
$\mathcal{L}= \alpha \mathcal{L}_{\rm bound} + \beta \mathcal{L}_{\rm  unbound}$
\cite{wang2019symmetric}, and (ii) designing bounded losses that decay slowly towards outliers, thus assigning them non-zero gradients \cite{ma2020normalized,zhou2021asymmetric,zhang2018generalized,amid2019robust}. In Tab.~\ref{Tab:LossFncts}, we show a comprehensive list of noise-robust loss functions that we will use to identify potential problems that arise when learning with these losses. The parameter settings for each of these loss functions, as per their respective publications, are documented in the appendix.

In Fig.~\ref{fig:learning_curves} we show the average error $\langle \delta_k \rangle$ in the ``correct'' neuron as a function of its pre-activation across a set of noise-robust loss functions.  Panel (a) shows a scenario with 10 classes for a neural network at initialization. The lower section of this panel represents the anticipated logit distribution $\mathcal{N}(0,1)$ for the neuron $k$.  It is obvious that these noise-robust loss functions deviate from the traditional Cross Entropy loss.  Specifically,  these functions assign gradients of reduced magnitude to examples with higher uncertainty (represented by small $z_k$) than the Cross Entropy loss. Despite this difference, the errors significantly deviate from zero at points of expected initial learning, as the histogram shows. 
However, in a 1000-class scenario, the overlap between the initial distribution and the regions where noise-robust losses record substantial errors is almost non-existent. Subsequent sections will show how this characteristic leads to negligible learning in the context of the 1000-class WebVision dataset.  Already in the case of 100 classes (not shown), the overlap is significantly reduced, slowing down learning on the Cifar-100 dataset.

Instead of using the unbounded Cross Entropy loss, which indiscriminately learns from outliers (indicated by very small  $z_k$ values), we suggest an alternative: adjusting the logit distribution to $p_z(z_k)=\mathcal{N}(\epsilon,1)$, 
to reestablish the crucial overlap with the error, as visually represented by the green histogram.

\subsection{Boosting learning with example-dependent logit bias}

In Fig.~\ref{fig:mae_error}, we have illustrated the challenge of using the MAE loss in scenarios with a large number of classes. The root of the challenge is the reduced overlap between the initial logit distribution and the region where the derivative of the loss is substantial. To tackle this, we introduce an example-dependent bias, denoted as $\epsilon$, to the logit, adjusting it from  $z_k$ to $z_k+\epsilon$. This adjustment ensures that the achieved learning speed is preserved irrespective of the class count. As before, $k$ refers to the class specified by the label.

The consequence of this adjustment for a newly initialized network is demonstrated in Fig.~\ref{fig:learning_curves}(b). Here, the logit distribution, $p_z(z_k)$, shifts from $\mathcal{N}(0,1)$ to $\mathcal{N}(\epsilon,1)$, reinstating the overlap with the region where significant gradients can emerge.   A noteworthy aspect of the MAE error is its implicit dependency on class count through the relation 
$\delta_n = 2(1- a_k)$. Since the  $a_k$   distribution is influenced by the class number, maintaining a consistent $a_k$ distribution would ensure a steady error.  To realize this, we aim to solve the following implicit equation for  $\epsilon(K)$:  
\begin{equation}
\left\langle \frac{\exp(z_k +\epsilon(K))}{\sum_{i=1}^K \exp(z_i +\delta[i-k] \epsilon(K) )}\right\rangle = C \ ,
\label{eq:constExpVal}
\end{equation}
where $C \in [0,1]$ represents a freely chosen constant which specifies the average activation of  $ a_k $ after the logit bias is applied.  
For the MAE loss our choice is $C =0.15$.  
This particular value ensures a substantial overlap between the network's initial logit distribution and the region where substantial gradients can occur, whilst excluding outliers with respect to the initial distribution.

We can evaluate  Eq.~\eqref{eq:constExpVal}  numerically for varying class counts. This is achieved by sampling Gaussian numbers, $z_i$, for a given $\epsilon$, and subsequently refining our estimate using techniques such as binary search. Results for the dependence of $\epsilon$ on the class number are presented in the appendix.
To clarify, when a logit bias is incorporated into a loss function, we append a ``*'' to the name of the loss.

\begin{figure}[t] %
  \centering
  \includegraphics[width=0.8\linewidth]{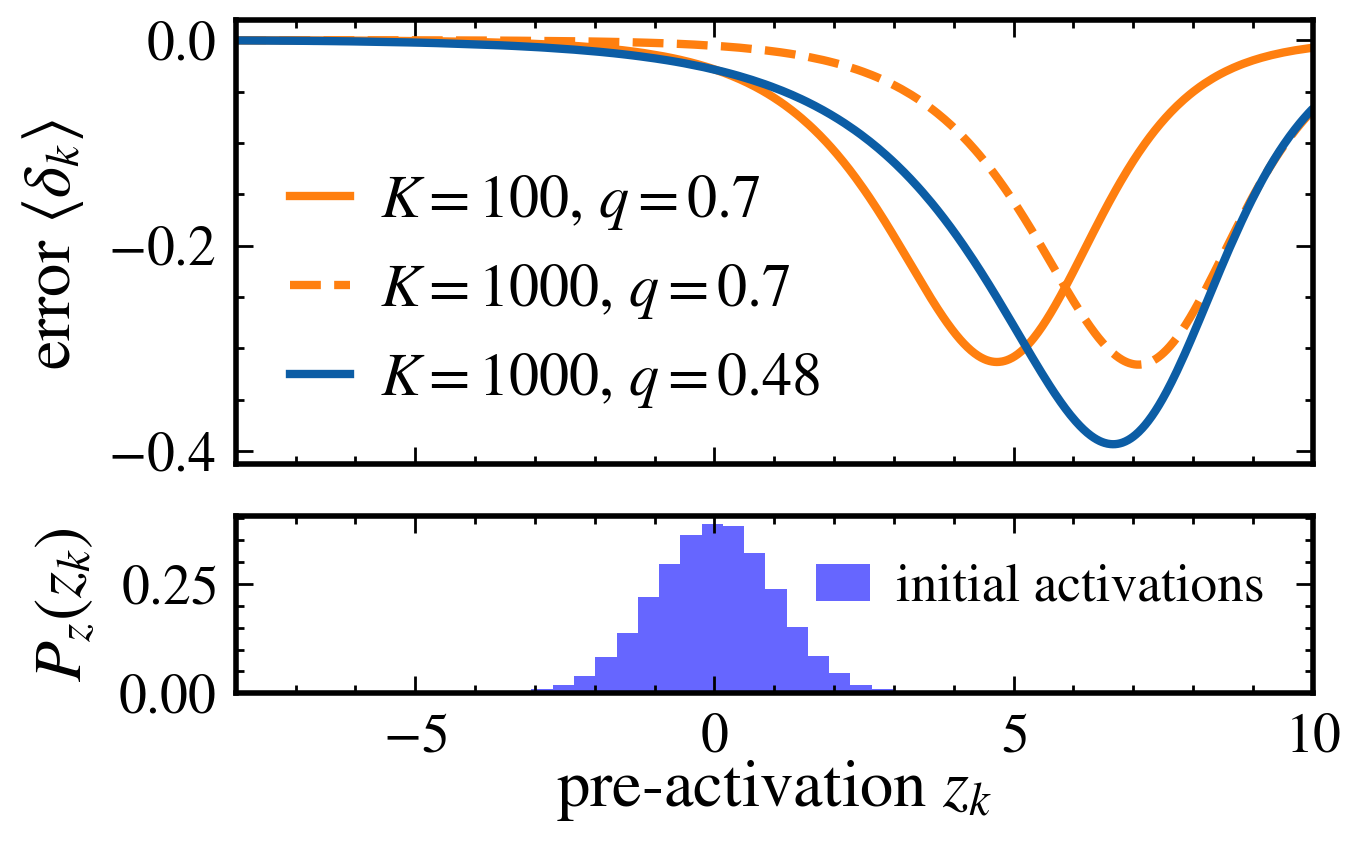}
  \caption{Computation of the hyperparameter $q$ for genCE when changing from $100$ to $1000$ classes. Instead of performing  
  an expensive hyperparameter search, we suggest to adjust the hyperparameter such that $\langle\delta_k \rangle$  at  position $z_k=0$ is unchanged for the larger class size.  }
    \label{fig:optimized_genCE}
  \end{figure}

\subsection{Calculation of hyperparameters \label{sec:calc_hyperpara}} 
An alternative approach is the adjustment of hyperparameters for different learning scenarios and datasets. However, choosing the right parameters can be highly non trivial and resource intensive. To reduce the number of hyperparameters, we propose to instead use the overlap shown in Fig.~\ref{fig:learning_curves} to  compute parameters without training.
Using hyperparameters empirically chosen for a
specific number of classes (e.g. for Cifar-10), one computes the value of the backpropagation error $\langle\delta_k\rangle (z_k)$ 
via  Eq.~\eqref{eq:back_prop_er} at the position $z_k = 0$ for this number of classes. When changing the number of classes, the hyperparameters are adjusted such that
$\langle\delta_k\rangle(0)$ coincides with the value of the baseline for which it is known that learning works well. In the case of a single unknown
this method yields the parameter as a function of the number of classes. In Fig.~\ref{fig:optimized_genCE} we demonstrate how this optimization changes the learning curve of genCE when going from $100$ to $1000$ classes. We find that the parameter $q=0.48$ provides the same $\langle \delta_k \rangle(0)$ for  $K=1000$ classes as in the case of $K=100$ with $q = 0.7$. This is empirically verified in Sec.~\ref{Sec:EmpRes} while further details on the calculation are shown in the appendix.

\section{Empirical Results}\label{Sec:EmpRes}

In the following, we present empirical evidence highlighting the efficiency of the logit bias in enhancing the learning capability of MAE on all datasets, demonstrating competitive or superior performance in the presence of  label noise.
In addition, we enable the loss functions genCE, NCE-AGCE and NF-MAE, to learn on the more complicated WebVision dataset using either new optimized hyperparameters or the logit bias, showcasing that the overlap between the initial output distribution and $\langle \delta_k \rangle$ is indeed the crucial requirement for learning. 
Datasets employed in this study include the publicly available datasets  Cifar-10, Cifar-100 \cite{Krizhevsky.2009}, Fashion-MNIST \cite{xiao2017fashion}, and the web-scraped WebVision dataset \cite{li2017webvision}. The noise profile in Cifar datasets is inherently minimal, enabling us to precisely adjust the label noise level. In contrast, WebVision naturally offers a more complex, potentially asymmetric noise structure.
ResNet architectures \cite{he2016deep}, tailored in size to the specific dataset, are used for most of our experiments.
We create symmetric label noise by assigning random labels to each example with probability $p$. In the case of asymmetric label noise we flip the labels of specific classes to different classes with probability $p$. The classes for the asymmetric noise are chosen identical to \cite{zhang2018generalized}.
Details on the training, together with further results from alternative network architectures, are available in the appendix.

Table~\ref{pred_table_2} summarizes the results on the datasets Cifar-10, Cifar-100 and FashionMNIST providing mean accuracies and associated errors (across five seeds) under various levels of label noise. The best performing loss functions are emphasized in bold for every combination of  network architecture and noise level.
For Cifar-10, we find that in the absence of label noise, MAE is unable to sufficiently learn the dataset, performing worse than all other loss functions. Yet, the introduction of a small
logit bias (specifically, $\epsilon=0.5$) in MAE* noticeably enhances the learning, bringing it on par with Cross Entropy loss.
In scenarios with symmetric label noise, MAE* maintains its noise resilience often performing best out of all loss functions.
Except for genCE and MAE*, most proposed loss functions either attain diminished accuracies in noise-free scenarios relative to the Cross Entropy or fit the  noise to a high degree. Overfitting is especially present for the Cross Entropy and biTemp loss, a trait anticipated from our theoretical discussions — especially noting biTemp's pronounced outlier tail in Fig.~\ref{fig:learning_curves}.
Similar results are found for the FashionMNIST dataset.

For the Cifar-100 dataset MAE* emerges as superior in the noiseless case, suggesting that even the minimal noise inherent in the Cifar-100 dataset \cite{northcutt2021pervasive} puts bounded loss functions in a favored position relative to CE. It is worth noting that $\epsilon$ is not optimized, but derived from Eq.~\eqref{eq:constExpVal}, based on the results of the ten-class tasks. 
For the case of Cifar-100  with
%
\begin{table*}[!h]
  \begin{tabular}{l l l l l l l l }
    \toprule 
    Dataset, Network   &Loss      & \multicolumn{1}{c}{Noise: 0 \%\,\,\,}    & \multicolumn{1}{c}{10 \%} & \multicolumn{1}{c}{20 \%}  & \multicolumn{1}{c}{40 \%} & \multicolumn{1}{c}{60 \%}  \\
    \midrule  
   \textbf{Sym. Noise}        &  CE                  & 91.30 $\pm$ 0.33          & 83.80 $\pm$ 0.36          & 78.21 $\pm$ 0.42          & 65.65 $\pm$ 0.23          & 50.12 $\pm$ 0.66         \\
          &  MAE                 & 84.90 $\pm$ 1.86          & 82.59 $\pm$ 1.22          & 80.45 $\pm$ 0.24          & 74.08 $\pm$ 4.17          & 59.90 $\pm$ 6.19         \\
          &  MAE* $\epsilon=0.5$  & 91.27 $\pm$ 0.13          & 90.67 $\pm$ 0.09          & \textbf{89.95} $\pm$ 0.06 & \textbf{87.33} $\pm$ 0.07 & 81.43 $\pm$ 0.19         \\
Cifar-10  & AGCE                & 88.32 $\pm$ 1.86          & 87.83 $\pm$ 1.87          & 87.06 $\pm$ 1.86          & 84.77 $\pm$ 1.76          & \textbf{82.24} $\pm$ 1.37         \\
          &  genCE               & 91.70 $\pm$ 0.10          & \textbf{91.12} $\pm$ 0.04 & 89.92 $\pm$ 0.11          & 86.81 $\pm$ 0.12          & 78.88 $\pm$ 0.30         \\
ResNet-32 & NCE-AGCE          & 64.70 $\pm$ 1.51          & 59.21 $\pm$ 1.13          & 55.77 $\pm$ 2.44          & 46.19 $\pm$ 3.56          & 35.76 $\pm$ 1.85\\
          &  NF-MAE             & 88.47 $\pm$ 0.26          & 87.64 $\pm$ 0.18          & 86.91 $\pm$ 0.07          & 84.05 $\pm$ 0.37          & 75.98 $\pm$ 0.49         \\
          &  NCE-MAE             & 88.57 $\pm$ 0.09          & 87.73 $\pm$ 0.14          & 86.81 $\pm$ 0.08          & 84.00 $\pm$ 0.19          & 75.67 $\pm$ 0.57         \\
          &  biTemp              & \textbf{92.07} $\pm$ 0.05 & 86.84 $\pm$ 0.17          & 81.33 $\pm$ 0.16          & 68.06 $\pm$ 0.63          & 51.58 $\pm$ 0.40         \\
\midrule
                  &  CE                  & \textbf{94.90} $\pm$ 0.07 & 92.41 $\pm$ 0.10          & 90.72 $\pm$ 0.07          & 85.60 $\pm$ 0.50          & 78.15 $\pm$ 0.71         \\
              &  MAE                 & 90.72 $\pm$ 3.37          & 93.74 $\pm$ 0.06          & 91.71 $\pm$ 1.88          & 88.24 $\pm$ 2.11          & 86.50 $\pm$ 2.16         \\
              &  MAE* $\epsilon=0.5$  & 94.77 $\pm$ 0.03          & \textbf{94.42} $\pm$ 0.02          & 94.23 $\pm$ 0.05          & \textbf{93.50} $\pm$ 0.11          & 92.04 $\pm$ 0.12\\
Fashion-MNIST & AGCE                & 94.39 $\pm$ 0.06          & 94.24 $\pm$ 0.07          & 94.01 $\pm$ 0.06          & 93.35 $\pm$ 0.06          & \textbf{92.28} $\pm$ 0.09\\
              &  genCE               & 94.79 $\pm$ 0.06          & 94.40 $\pm$ 0.06          & \textbf{94.28} $\pm$ 0.02 & 93.46 $\pm$ 0.07          & 91.11 $\pm$ 0.15         \\
ResNet-32     & NCE-AGCE          & 92.73 $\pm$ 0.10          & 92.57 $\pm$ 0.10          & 92.12 $\pm$ 0.05          & 91.58 $\pm$ 0.09          & 90.09 $\pm$ 0.13         \\
              &  NF-MAE             & 93.92 $\pm$ 0.07          & 93.74 $\pm$ 0.06          & 93.71 $\pm$ 0.06          & 92.84 $\pm$ 0.07          & 91.75 $\pm$ 0.10         \\
              &  NCE-MAE             & 93.98 $\pm$ 0.04          & 93.73 $\pm$ 0.07          & 93.51 $\pm$ 0.06          & 93.00 $\pm$ 0.07          & 91.73 $\pm$ 0.12         \\
              &  biTemp              & 94.78 $\pm$ 0.03          & 93.38 $\pm$ 0.03          & 91.49 $\pm$ 0.09          & 86.65 $\pm$ 0.50          & 78.05 $\pm$ 0.47         \\
\midrule
     &  CE                       & 75.88 $\pm$ 0.15          & 69.21 $\pm$ 0.12          & 63.07 $\pm$ 0.15          & 46.46 $\pm$ 0.24          & 24.69 $\pm$ 0.68         \\
             & MAE                  & \phantom{0}6.84 $\pm$ 0.66           & \phantom{0}5.30 $\pm$ 0.36           & \phantom{0}4.93 $\pm$ 0.43           & \phantom{0}2.54 $\pm$ 0.29           & \phantom{0}1.67 $\pm$ 0.17          \\
          &  MAE* $\epsilon=1.5$      & 74.81 $\pm$ 0.13          & 72.33 $\pm$ 0.13          & 68.30 $\pm$ 0.24          & 55.71 $\pm$ 0.45          & 35.51 $\pm$ 0.19         \\
          &  MAE* $\epsilon=3.0$      & \textbf{76.49} $\pm$ 0.09          & 69.11 $\pm$ 0.14          & 61.13 $\pm$ 0.16          & 44.20 $\pm$ 0.41          & 26.45 $\pm$ 0.16         \\
Cifar-100 &  AGCE                 & 76.32 $\pm$ 0.12          & \textbf{73.23} $\pm$ 0.11 & 67.91 $\pm$ 0.19          & 52.64 $\pm$ 0.15          & 32.48 $\pm$ 0.12         \\         
 &  genCE                    & 73.90 $\pm$ 0.17          & 72.15 $\pm$ 0.09          & 69.37 $\pm$ 0.20          & \textbf{61.15} $\pm$ 0.10 & \textbf{41.52} $\pm$ 0.64\\
 ResNet-34         & NCE-AGCE               & 72.46 $\pm$ 0.09          & 70.00 $\pm$ 0.30          & 66.20 $\pm$ 0.35          & 46.91 $\pm$ 0.69          & 10.86 $\pm$ 0.42   \\     
         &  NF-MAE                  & 74.66 $\pm$ 0.17          & 72.91  $\pm$ 0.14 & \textbf{70.25} $\pm$ 0.06 & 60.97 $\pm$ 0.14          & 25.23 $\pm$ 0.68         \\
          &  NCE-MAE                  & 74.85 $\pm$ 0.10          &  72.71  $\pm$ 0.07          & 70.03 $\pm$ 0.10          & 60.51 $\pm$ 0.10          & 26.77 $\pm$ 0.81         \\
          &  biTemp                   & 76.19 $\pm$ 0.14          & 68.40 $\pm$ 0.12          & 61.36 $\pm$ 0.19          & 46.15 $\pm$ 0.23          & 26.67 $\pm$ 0.16         \\
\midrule
 \textbf{Asym. Noise}   & CE                  & 91.98 $\pm$ 0.11          & 91.62 $\pm$ 0.20          & 91.63 $\pm$ 0.27          & 85.28 $\pm$ 6.82          & 64.86 $\pm$ 6.71         \\
          & MAE                 & 83.41 $\pm$ 1.49          & 83.62 $\pm$ 1.48          & 80.02 $\pm$ 1.79          & 77.25 $\pm$ 6.32          & 65.35 $\pm$ 6.40         \\
         & MAE* $\epsilon=0.5$  & 91.14 $\pm$ 0.08          & 91.15 $\pm$ 0.09          & 91.33 $\pm$ 0.15          & \textbf{85.82} $\pm$ 5.68 & 65.51 $\pm$ 6.48         \\
Cifar-10   & AGCE                & 89.45 $\pm$ 1.57          & 87.80 $\pm$ 1.87          & 89.43 $\pm$ 1.54          & 83.78 $\pm$ 5.51          & \textbf{68.40} $\pm$ 5.63\\
           & genCE               & 91.84 $\pm$ 0.06          & 91.65 $\pm$ 0.06          & 91.69 $\pm$ 0.10          & 84.86 $\pm$ 6.74          & 64.51 $\pm$ 6.72         \\
ResNet-32  & NCE-AGCE            & 61.21 $\pm$ 1.26          & 64.00 $\pm$ 1.97          & 62.74 $\pm$ 1.65          & 55.29 $\pm$ 2.94          & 49.10 $\pm$ 2.96         \\
           & NF-MAE              & 88.50 $\pm$ 0.05          & 88.49 $\pm$ 0.17          & 88.71 $\pm$ 0.07          & 82.03 $\pm$ 6.29          & 63.12 $\pm$ 6.31         \\
           & NCE-MAE             & 88.60 $\pm$ 0.11          & 88.41 $\pm$ 0.08          & 88.40 $\pm$ 0.15          & 82.22 $\pm$ 6.40          & 63.03 $\pm$ 6.27         \\
           & biTemp              & \textbf{92.24} $\pm$ 0.04 & \textbf{92.19} $\pm$ 0.08 & \textbf{92.27} $\pm$ 0.06 & 85.52 $\pm$ 6.86          & 65.07 $\pm$ 6.78         \\
\midrule
                       & CE                  & \textbf{94.91} $\pm$ 0.07 & 92.38 $\pm$ 0.13          & 90.38 $\pm$ 0.16          & 85.61 $\pm$ 0.20          & 78.07 $\pm$ 0.72         \\
           & MAE                 & 91.94 $\pm$ 1.84          & 93.81 $\pm$ 0.10          & 93.38 $\pm$ 0.16          & 87.91 $\pm$ 3.26          & 88.83 $\pm$ 1.85         \\
           & MAE* $\epsilon=0.5$  & 94.67 $\pm$ 0.08          & \textbf{94.48} $\pm$ 0.06 & 94.19 $\pm$ 0.06          & \textbf{93.33} $\pm$ 0.04 & 91.86 $\pm$ 0.04         \\
FashionMNIST   & AGCE                & 94.44 $\pm$ 0.12          & 94.31 $\pm$ 0.11          & 94.01 $\pm$ 0.07          & 93.31 $\pm$ 0.07          & \textbf{92.25} $\pm$ 0.10\\
           & genCE               & 94.81 $\pm$ 0.06          & 94.47 $\pm$ 0.03          & \textbf{94.23} $\pm$ 0.09 & 93.27 $\pm$ 0.03          & 90.75 $\pm$ 0.15         \\
ResNet-32  & NCE-AGCE            & 92.69 $\pm$ 0.04          & 92.63 $\pm$ 0.08          & 92.28 $\pm$ 0.17          & 91.71 $\pm$ 0.09          & 90.42 $\pm$ 0.17         \\
           & NF-MAE              & 93.92 $\pm$ 0.06          & 93.81 $\pm$ 0.06          & 93.55 $\pm$ 0.04          & 92.82 $\pm$ 0.10          & 91.75 $\pm$ 0.17         \\
           & NCE-MAE             & 94.07 $\pm$ 0.05          & 93.81 $\pm$ 0.06          & 93.57 $\pm$ 0.05          & 92.93 $\pm$ 0.05          & 91.74 $\pm$ 0.06         \\
           & biTemp              & 94.90 $\pm$ 0.08          & 93.37 $\pm$ 0.10          & 91.47 $\pm$ 0.11          & 86.64 $\pm$ 0.34          & 77.56 $\pm$ 0.62         \\
    \bottomrule   
  \end{tabular}
  \caption{Final test accuracies when training with different loss functions on the Cifar-10, Fashion-MNIST and Cifar-100 dataset using various amounts of label noise. In the case of symmetric label noise with ten-classes (first two panels) we find that MAE* improves significantly over MAE, performing as well or superior to the CE in the noiseless case while often achieving the top result in the case of label noise. For the Cifar-100 dataset, the logit bias allows MAE* to learn exceptionally well compared to MAE, even achieving the best test-accuracy under all loss functions on the pristine dataset.
  For larger label noises MAE* with the calculated logit bias $\epsilon=3.0$ overfits, showcasing that a smaller logit bias (e.g. $\epsilon=1.5$) might be preferable. In the case of asymmetric noise (bottom panels) MAE* again improves significantly over MAE on all noise level. When compared to the other loss functions we find that counted over all datasets and noise levels MAE* achieves the most top accuracies.} 
  \label{pred_table_2} 
\end{table*}  
\begin{table*}[!ht]
\centering
  \begin{tabular}{l c c c c c c c}
    \toprule 
      & NF-MAE  & NCE-AGCE  & genCE & CE    &  biTemp  & MAE*   \\
    \midrule
    Top-1 & 1.04 & 7.48 & 13.28 & 62.55 $\pm$ 0.16  & 63.60 $\pm$ 0.14 & 58.72 $\pm$ 0.11   \\
    Top-5 & 1.17 & 10.08 & 15.39 & 79.63 $\pm$ 0.10   & 80.66 $\pm$ 0.13 & 79.68 $\pm$ 0.04  \\
    \midrule
    & $\phantom{*}$NF-MAE** & $\phantom{+}$NCE-AGCE+ & genCE** &  genCE+  & MAE**\\
    \midrule
    Top-1 & 62.30 $\pm$ 0.01 & 63.90 $\pm$ 0.09 & 62.03 $\pm$ 0.07  & \textbf{64.85} $\pm$ 0.04 & 62.70 $\pm$ 0.07 \\
     Top-5 & 80.95 $\pm$ 0.07 & 80.22 $\pm$ 0.09  & 80.70  $\pm$ 0.03 & 80.93 $\pm$ 0.02 & \textbf{81.11} $\pm$ 0.05 \\
    \bottomrule 
  \end{tabular}
   \caption{Final accuracies on the validation images of the 1000-class WebVision dataset after training Wide-ResNet-101 with different loss functions. If an error is provided, accuracies for five different seeds are averaged. For the losses that do not learn, we report the accuracies of a single training run. The upper row demonstrates that all loss function which where noise robust on Cifar-10 and Cifar-100 are unable to learn on WebVision, with the exception of MAE*. The second row demonstrates that adding either a scheduled logit bias (indicated with **) or calculating new parameters for the loss as described by our theory (denoted by +), all loss functions are capable of learning the WebVision dataset. The exact parameters in the second row are provided in the appendix.}
  \label{pred_table_WebVis}
\end{table*} 
added noise, a different picture emerges: The loss functions CE, biTemp, and MAE* tend to fit the noise, while the loss functions which underfitted the pristine dataset now display superior generalisation. This is especially pronounced for genCE, which lacks $2.5\%$ behind in the noiseless case but performs best in the case of larger noise levels. 
For demonstration purpose we also add  $\epsilon=1.5$ to the table, showcasing that the logit bias $\epsilon$ can be used to tune a balance between learning speed and noise-robustness. 
The case of asymmetric noise is summarized in the bottom two panels
of Tab.~\ref{pred_table_2}. In the case of MAE* we find very similar results to the case of symmetric label noise in terms of improvements over MAE, as well as in cases where it performs best. Considering all datasets and noise variants, MAE* with logit bias is the loss function with the most top accuracies.

In the 100 class scenario discussed above  the majority of the studied loss functions still demonstrate good performance, albeit at a marginally reduced learning speed.
Their design incorporates ``tails''  that encompass the logit distribution at initialization.  These tails ensure non-zero gradients, enabling the learning of the dataset. However, when scaling to 1000 classes, most tails appear to fall flat. This is demonstrated in the first row of Tab.~\ref{pred_table_WebVis}, where we find that all loss function which where noise robust on Cifar-10 and Cifar-100 are unable to learn on the WebVision dataset, with the exception of MAE*. The second row demonstrates that if we restore the overlap with the initial distribution using either the logit bias or by calculating new parameters for these losses as described in Sec.~\ref{sec:calc_hyperpara} (denoted by +), these losses are now capable of learning on the WebVision dataset. 

Though learning works well with the calculated value of the logit bias $\epsilon=5.2$, removal of the logit bias during testing reveals reduced confidence in the correct label, induced by the elevated $\epsilon$ value during the training phase. This can be mitigated by weaning the network from its high confidence prediction by adopting smaller values of $\epsilon$ throughout training using an $\epsilon$ schedule (denoted by **), as elaborated in the appendix. Comparing the losses against each other, we find that the $\epsilon$ enhanced losses NF-MAE**, genCE**, and MAE** perform best in the top-5
accuracy, while 
genCE+ and NCE-AGCE+ with calculated new hyperparameters perform best in the top-1 accuracy, showcasing that our method provides the missing ingredient to these losses.

\emph{Limitations}---
Our theoretical insights indicate that our primary findings should exhibit general applicability across numerous learning tasks, however, the current scope is confined  to image classification. 
Furthermore, it seems that our present computation of $\epsilon$ based on the number of classes tends to overestimate the required $\epsilon$, which can compromise noise robustness. To address this issue, future theoretical studies could explore a more sophisticated understanding of this phenomenon, while empirical analyses could configure $\epsilon$ as a tunable hyperparameter to circumvent this limitation.

\section{Conclusion } 

In this study, we analyzed the dynamics of early-stage learning by computing the average backpropagation error, offering quantitative insights into how increasing the class count influences initial learning, especially in the context of bounded loss functions. Capitalizing on these insights, we showed how learning with bounded losses can be enhanced by calculating how specific hyperparameters depend on the number of classes, effectively eradicating the need of extensive hyperparameter searching. For the hyperparameter-free case of the MAE-loss, we enabled learning using the \emph{logit bias}. Both techniques use the method of realigning the distribution of a newly initialized network with the region where the loss function has a nonzero derivative, enabling effective learning even in scenarios with a multitude of classes. 
Empirical evidence underscores the effectiveness of this method. The loss functions with calculated hyperparameters emerged as the best performing loss functions on WebVision regarding the top-1 accuracy, while the logit bias enhanced MAE loss demonstrates comparable, if not superior, performance across datasets spanning ten to a thousand classes. Before our approach, such outcomes were largely exclusive to the Cross Entropy or biTemp loss. However, both these methods tend to overfit.
In summary, we argue that our method is a first step towards a comprehensive framework -- one that allows for noise robust learning, regardless of the number of classes, and without an over-reliance on fine-tuned hyperparameters.

\section*{Acknowledgments}
Computations for this work were done (in part) using resources of the Leipzig University Computing Center.

\bibliography{aaai25}

\begin{thebibliography}{43}
\providecommand{\natexlab}[1]{#1}

\bibitem[{Amid et~al.(2019)Amid, Warmuth, Anil, and Koren}]{amid2019robust}
Amid, E.; Warmuth, M.~K.; Anil, R.; and Koren, T. 2019.
\newblock Robust bi-tempered logistic loss based on bregman divergences.
\newblock \emph{Advances in Neural Information Processing Systems}, 32.

\bibitem[{Author(s)(2023)}]{code}
Author(s), A. 2023.
\newblock {All code, scripts, and data used in this work are included in a Zenodo archive: \url{https://zenodo.org/records/13150928}}.
\newblock \emph{Zenodo}.

\bibitem[{Chen et~al.(2020)Chen, Kornblith, Norouzi, and Hinton}]{chen2020simple}
Chen, T.; Kornblith, S.; Norouzi, M.; and Hinton, G. 2020.
\newblock A simple framework for contrastive learning of visual representations.
\newblock In \emph{International conference on machine learning}, 1597--1607. PMLR.

\bibitem[{Englesson and Azizpour(2021)}]{englesson2021generalized}
Englesson, E.; and Azizpour, H. 2021.
\newblock Generalized jensen-shannon divergence loss for learning with noisy labels.
\newblock \emph{Advances in Neural Information Processing Systems}, 34: 30284--30297.

\bibitem[{Feng et~al.(2021)Feng, Shu, Lin, Lv, Li, and An}]{feng2021can}
Feng, L.; Shu, S.; Lin, Z.; Lv, F.; Li, L.; and An, B. 2021.
\newblock Can cross entropy loss be robust to label noise?
\newblock In \emph{Proceedings of the Twenty-Ninth International Conference on International Joint Conferences on Artificial Intelligence}, 2206--2212.

\bibitem[{Ghosh, Kumar, and Sastry(2017)}]{ghosh2017robust}
Ghosh, A.; Kumar, H.; and Sastry, P.~S. 2017.
\newblock Robust loss functions under label noise for deep neural networks.
\newblock In \emph{Proceedings of the AAAI conference on artificial intelligence}, volume~31.

\bibitem[{Ghosh and Lan(2021)}]{ghosh2021contrastive}
Ghosh, A.; and Lan, A. 2021.
\newblock Contrastive learning improves model robustness under label noise.
\newblock In \emph{Proceedings of the IEEE/CVF Conference on Computer Vision and Pattern Recognition}, 2703--2708.

\bibitem[{Goldberger and Ben-Reuven(2017)}]{goldberger2017training}
Goldberger, J.; and Ben-Reuven, E. 2017.
\newblock Training deep neural-networks using a noise adaptation layer.
\newblock In \emph{International conference on learning representations}.

\bibitem[{Goodfellow, Bengio, and Courville(2016)}]{goodfellow2016deep}
Goodfellow, I.; Bengio, Y.; and Courville, A. 2016.
\newblock \emph{Deep learning}.
\newblock MIT press.

\bibitem[{Han et~al.(2018)Han, Yao, Niu, Zhou, Tsang, Zhang, and Sugiyama}]{han2018masking}
Han, B.; Yao, J.; Niu, G.; Zhou, M.; Tsang, I.; Zhang, Y.; and Sugiyama, M. 2018.
\newblock Masking: A new perspective of noisy supervision.
\newblock \emph{Advances in neural information processing systems}, 31.

\bibitem[{He et~al.(2015)He, Zhang, Ren, and Sun}]{he2015delving}
He, K.; Zhang, X.; Ren, S.; and Sun, J. 2015.
\newblock Delving deep into rectifiers: Surpassing human-level performance on imagenet classification.
\newblock In \emph{Proceedings of the IEEE international conference on computer vision}, 1026--1034.

\bibitem[{He et~al.(2016)He, Zhang, Ren, and Sun}]{he2016deep}
He, K.; Zhang, X.; Ren, S.; and Sun, J. 2016.
\newblock Deep residual learning for image recognition.
\newblock In \emph{Proceedings of the IEEE conference on computer vision and pattern recognition}, 770--778.

\bibitem[{Hendrycks et~al.(2019)Hendrycks, Mazeika, Kadavath, and Song}]{hendrycks2019using}
Hendrycks, D.; Mazeika, M.; Kadavath, S.; and Song, D. 2019.
\newblock Using self-supervised learning can improve model robustness and uncertainty.
\newblock \emph{Advances in neural information processing systems}, 32.

\bibitem[{Idelbayev(2020)}]{Idelbayev2020}
Idelbayev, Y. 2020.
\newblock Pytorch ResNet Cifar-10.
\newblock \url{https://github.com/akamaster/pytorch_resnet_cifar10/blob/master/trainer.py}.

\bibitem[{Khetan, Lipton, and Anandkumar(2017)}]{khetan2017learning}
Khetan, A.; Lipton, Z.~C.; and Anandkumar, A. 2017.
\newblock Learning from noisy singly-labeled data.
\newblock \emph{arXiv preprint arXiv:1712.04577}.

\bibitem[{Krizhevsky, Hinton et~al.(2009)}]{Krizhevsky.2009}
Krizhevsky, A.; Hinton, G.; et~al. 2009.
\newblock Learning multiple layers of features from tiny images.
\newblock \emph{{Tech Report}}.

\bibitem[{Lee et~al.(2017)Lee, Bahri, Novak, Schoenholz, Pennington, and Sohl-Dickstein}]{lee2017deep}
Lee, J.; Bahri, Y.; Novak, R.; Schoenholz, S.~S.; Pennington, J.; and Sohl-Dickstein, J. 2017.
\newblock Deep neural networks as gaussian processes.
\newblock \emph{arXiv preprint arXiv:1711.00165}.

\bibitem[{Lee et~al.(2018)Lee, He, Zhang, and Yang}]{lee2018cleannet}
Lee, K.-H.; He, X.; Zhang, L.; and Yang, L. 2018.
\newblock Cleannet: Transfer learning for scalable image classifier training with label noise.
\newblock In \emph{Proceedings of the IEEE conference on computer vision and pattern recognition}, 5447--5456.

\bibitem[{Li et~al.(2017)Li, Wang, Li, Agustsson, and Van~Gool}]{li2017webvision}
Li, W.; Wang, L.; Li, W.; Agustsson, E.; and Van~Gool, L. 2017.
\newblock Webvision database: Visual learning and understanding from web data.
\newblock \emph{arXiv preprint arXiv:1708.02862}.

\bibitem[{Liang, Liu, and Yao(2022)}]{liang2022review}
Liang, X.; Liu, X.; and Yao, L. 2022.
\newblock Review--a survey of learning from noisy labels.
\newblock \emph{ECS Sensors Plus}, 1(2): 021401.

\bibitem[{Lyu and Tsang(2019)}]{lyu2019curriculum}
Lyu, Y.; and Tsang, I.~W. 2019.
\newblock Curriculum loss: Robust learning and generalization against label corruption.
\newblock \emph{arXiv preprint arXiv:1905.10045}.

\bibitem[{Ma et~al.(2020)Ma, Huang, Wang, Romano, Erfani, and Bailey}]{ma2020normalized}
Ma, X.; Huang, H.; Wang, Y.; Romano, S.; Erfani, S.; and Bailey, J. 2020.
\newblock Normalized loss functions for deep learning with noisy labels.
\newblock In \emph{International conference on machine learning}, 6543--6553. PMLR.

\bibitem[{Neal(1994)}]{neal1994priors}
Neal, R.~M. 1994.
\newblock Priors for infinite networks (tech. rep. no. crg-tr-94-1).
\newblock \emph{University of Toronto}, 415.

\bibitem[{Northcutt, Athalye, and Mueller(2021)}]{northcutt2021pervasive}
Northcutt, C.~G.; Athalye, A.; and Mueller, J. 2021.
\newblock Pervasive label errors in test sets destabilize machine learning benchmarks.
\newblock \emph{arXiv preprint arXiv:2103.14749}.

\bibitem[{Reed et~al.(2014)Reed, Lee, Anguelov, Szegedy, Erhan, and Rabinovich}]{reed2014training}
Reed, S.; Lee, H.; Anguelov, D.; Szegedy, C.; Erhan, D.; and Rabinovich, A. 2014.
\newblock Training deep neural networks on noisy labels with bootstrapping.
\newblock \emph{arXiv preprint arXiv:1412.6596}.

\bibitem[{Ren et~al.(2018)Ren, Zeng, Yang, and Urtasun}]{ren2018learning}
Ren, M.; Zeng, W.; Yang, B.; and Urtasun, R. 2018.
\newblock Learning to reweight examples for robust deep learning.
\newblock In \emph{International conference on machine learning}, 4334--4343. PMLR.

\bibitem[{Song et~al.(2022)Song, Kim, Park, Shin, and Lee}]{song2022learning}
Song, H.; Kim, M.; Park, D.; Shin, Y.; and Lee, J.-G. 2022.
\newblock Learning from noisy labels with deep neural networks: A survey.
\newblock \emph{IEEE Transactions on Neural Networks and Learning Systems}.

\bibitem[{Sukhbaatar and Fergus(2014)}]{sukhbaatar2014learning}
Sukhbaatar, S.; and Fergus, R. 2014.
\newblock Learning from noisy labels with deep neural networks.
\newblock \emph{arXiv preprint arXiv:1406.2080}, 2(3): 4.

\bibitem[{Thulasidasan et~al.(2019)Thulasidasan, Bhattacharya, Bilmes, Chennupati, and Mohd-Yusof}]{thulasidasan2019combating}
Thulasidasan, S.; Bhattacharya, T.; Bilmes, J.; Chennupati, G.; and Mohd-Yusof, J. 2019.
\newblock Combating label noise in deep learning using abstention.
\newblock \emph{arXiv preprint arXiv:1905.10964}.

\bibitem[{Toner and Storkey(2023)}]{toner2023label}
Toner, W.; and Storkey, A. 2023.
\newblock Label Noise: Correcting a Correction.
\newblock \emph{arXiv preprint arXiv:2307.13100}.

\bibitem[{Veit et~al.(2017)Veit, Alldrin, Chechik, Krasin, Gupta, and Belongie}]{veit2017learning}
Veit, A.; Alldrin, N.; Chechik, G.; Krasin, I.; Gupta, A.; and Belongie, S. 2017.
\newblock Learning from noisy large-scale datasets with minimal supervision.
\newblock In \emph{Proceedings of the IEEE conference on computer vision and pattern recognition}, 839--847.

\bibitem[{Wang et~al.(2019)Wang, Ma, Chen, Luo, Yi, and Bailey}]{wang2019symmetric}
Wang, Y.; Ma, X.; Chen, Z.; Luo, Y.; Yi, J.; and Bailey, J. 2019.
\newblock Symmetric cross entropy for robust learning with noisy labels.
\newblock In \emph{Proceedings of the IEEE/CVF International Conference on Computer Vision}, 322--330.

\bibitem[{weiaicunzai(2020)}]{weiaicunzai2020}
weiaicunzai. 2020.
\newblock Pytorch Cifar-100.
\newblock \url{https://github.com/weiaicunzai/pytorch-cifar100/blob/master/models/resnet.py}.

\bibitem[{{Xavier Glorot} and {Yoshua Bengio}(2010)}]{Glorot.2010}
{Xavier Glorot}; and {Yoshua Bengio}. 2010.
\newblock {Understanding the difficulty of training deep feedforward neural networks}.
\newblock \emph{{Proceedings of the Thirteenth International Conference on Artificial Intelligence and Statistics}}, 249--256.

\bibitem[{Xiao, Rasul, and Vollgraf(2017)}]{xiao2017fashion}
Xiao, H.; Rasul, K.; and Vollgraf, R. 2017.
\newblock Fashion-mnist: a novel image dataset for benchmarking machine learning algorithms.
\newblock \emph{arXiv preprint arXiv:1708.07747}.

\bibitem[{Xiao et~al.(2015)Xiao, Xia, Yang, Huang, and Wang}]{xiao2015learning}
Xiao, T.; Xia, T.; Yang, Y.; Huang, C.; and Wang, X. 2015.
\newblock Learning from massive noisy labeled data for image classification.
\newblock In \emph{Proceedings of the IEEE conference on computer vision and pattern recognition}, 2691--2699.

\bibitem[{Xu et~al.(2019)Xu, Cao, Kong, and Wang}]{xu2019l_dmi}
Xu, Y.; Cao, P.; Kong, Y.; and Wang, Y. 2019.
\newblock L\_dmi: A novel information-theoretic loss function for training deep nets robust to label noise.
\newblock \emph{Advances in neural information processing systems}, 32.

\bibitem[{Xue, Whitecross, and Mirzasoleiman(2022)}]{xue2022investigating}
Xue, Y.; Whitecross, K.; and Mirzasoleiman, B. 2022.
\newblock Investigating why contrastive learning benefits robustness against label noise.
\newblock In \emph{International Conference on Machine Learning}, 24851--24871. PMLR.

\bibitem[{Yang et~al.(2021)Yang, Hu, Babuschkin, Sidor, Liu, Farhi, Ryder, Pachocki, Chen, and Gao}]{yang2021tuning}
Yang, G.; Hu, E.; Babuschkin, I.; Sidor, S.; Liu, X.; Farhi, D.; Ryder, N.; Pachocki, J.; Chen, W.; and Gao, J. 2021.
\newblock Tuning large neural networks via zero-shot hyperparameter transfer.
\newblock \emph{Advances in Neural Information Processing Systems}, 34: 17084--17097.

\bibitem[{Yi et~al.(2022)Yi, Liu, She, McLeod, and Wang}]{yi2022learning}
Yi, L.; Liu, S.; She, Q.; McLeod, A.~I.; and Wang, B. 2022.
\newblock On learning contrastive representations for learning with noisy labels.
\newblock In \emph{Proceedings of the IEEE/CVF conference on computer vision and pattern recognition}, 16682--16691.

\bibitem[{Zhang and Sabuncu(2018)}]{zhang2018generalized}
Zhang, Z.; and Sabuncu, M. 2018.
\newblock Generalized cross entropy loss for training deep neural networks with noisy labels.
\newblock \emph{Advances in neural information processing systems}, 31.

\bibitem[{Zheltonozhskii et~al.(2022)Zheltonozhskii, Baskin, Mendelson, Bronstein, and Litany}]{zheltonozhskii2022}
Zheltonozhskii, E.; Baskin, C.; Mendelson, A.; Bronstein, A.~M.; and Litany, O. 2022.
\newblock Contrast to divide: Self-supervised pre-training for learning with noisy labels.
\newblock In \emph{Proceedings of the IEEE/CVF Winter Conference on Applications of Computer Vision}, 1657--1667.

\bibitem[{Zhou et~al.(2021)Zhou, Liu, Jiang, Gao, and Ji}]{zhou2021asymmetric}
Zhou, X.; Liu, X.; Jiang, J.; Gao, X.; and Ji, X. 2021.
\newblock Asymmetric loss functions for learning with noisy labels.
\newblock In \emph{International conference on machine learning}, 12846--12856. PMLR.

\end{thebibliography}

\newpage 
~
\newpage

\appendix

\section{Appendix}

\begin{table*}[t]
\centering
  \begin{tabular}{l l l l l  }
 \toprule 
    & Fashion-MNIST & Cifar-10 & Cifar-100 & WebVision \\
   biTemp & $t_1=0.8$, $t_2=1.2 $ & $t_1=0.8$, $t_2=1.2 $ & $t_1=0.8$, $t_2=1.2 $ & $t_1=0.8$, $t_2=1.2 $  \\
   genCE & $q=0.7$ & $q=0.7$ & $q=0.7$ & $q=0.7$ \\
    AGCE & $a=0.6$, $q=0.6$ & $a=0.6$, $q=0.6$ & $a=1e-5$, $q=0.5$ & $a=1e-5$, $q=0.5$  \\
    NF-MAE & $\alpha=1$, $\beta=20$  & $\alpha=1$, $\beta=20$  & $\alpha=1$,$\beta=0.2$  & $\alpha=1$,$\beta=0.2$   \\
    NCE-MAE & $\alpha=1$, $\beta=20$  & $\alpha=1$, $\beta=20$  & $\alpha=1$, $\beta=20$  & $\alpha=1$, $\beta=20$   \\
    \multirow{ 2}{*}{NCE-AGCE} & $\alpha=1$, $\beta=4$  & $\alpha=1$, $\beta=4$  & $\alpha=10$, $\beta=0.1$  & $\alpha=10$, $\beta=0.1$   \\
    & $a=6$, $q=1.5$ & $a=6$, $q=1.5$ & $a=1.8$, $q=3$ & $a=1.8$, $q=3$  \\
    \bottomrule   
  \end{tabular}
   \caption{Hyperparameters for the different loss functions on various datasets. The parameters are adapted from the respective papers, in the case of WebVision we took the parameters used for Cifar-100. }
  \label{tab:hyperP}
\end{table*}  

\subsection{Training Parameters \label{Apend:training_p}}

Here, we describe the experimental framework used to compare various loss functions.

\emph{Networks}---We train two distinct types of network architectures: (i) residual, convolutional networks utilizing the ResNet architecture \citep{he2016deep} with ReLU activation\footnote{For ResNet networks, we adhere to the \texttt{pytorch} implementation described in Refs.~\citet{Idelbayev2020,weiaicunzai2020} for Cifar-10 and 100 respectively, while Wide-ResNet-101 is used directly from the \texttt{pytorch} library.}, and (ii) multilayer feedforward neural networks, termed MLP1024, comprised of dataset-specific input and output layers, along with three hidden layers of sizes 1024, 512, and 512, also with ReLU activations. The results of the MLP1024 network are only shown in the appendix.

\emph{Preprocessing}---When training ResNet networks, we normalize Cifar-10 images by subtracting the mean $\mu=(0.485,0.456,0.406)$ and then dividing by $\sigma=(0.229,0.224,0.225)$ for the three color channels. On each training data batch, we further perform a random horizontal flip followed by random cropping to size $32\times32$ with a padding of size $4$. 

For training ResNet on Fashion-MNIST, we normalize the black and white images using $\mu=0.286$ and $\sigma=0.353$. Batches during training are processed with a random horizontal flip followed by random cropping to a size $28\times28$ with a padding of size $4$.
The Cifar-100 dataset is normalized using $\mu=(0.507,0.487,0.441)$ and $\sigma=(0.267,0.256,0.276)$ and the training data is processed with random cropping ($32\times32$, padding $4$), followed by a random horizontal flip, and a random rotation by up to $15^\circ$.
Training ResNet networks on WebVision, we normalize the images by subtracting the mean $\mu=(0.485, 0.456, 0.406)$ and then dividing by $\sigma=(0.229, 0.224, 0.225)$ for the three color channels. On each training data batch, we further perform a random horizontal flip followed by random cropping to size $224\times224$. 

For training MLP1024 networks, we only normalize the images. All test data is normalized identically to the corresponding training data and no further processing is performed.

\emph{Training}---All ResNet networks, are trained using stochastic gradient descent (SGD) with a momentum of $0.9$, a weight decay of $10^{-4}$, and a step-wise learning rate schedule with an initial learning rate of $0.1$. On Fashion-MNIST and Cifar-10 the learning rate is multiplied with a factor of $0.1$ at epochs $100$ and $150$. For Cifar-100 the learning rate changes at epochs $60,120,160$ by a factor of $0.2$ and for WebVision we change at epochs $66$ and $132$ by a factor of $0.1$. The mini-batch size is $128$ in all cases but WebVision, where we change it to $400$. For WebVision, we further reduced the training time by only considering the Google images, reducing the number of images from $2.4$ million to a little more than a million.

MLP1024 networks are trained with SGD, with momentum set to $0.95$, a mini-batch size of $32$, and no weight decay. We use an exponential learning rate schedule with a decay factor of $0.95$ for each epoch. For the small MLP1024, we additionally perform a grid searches over the initial learning rates.

The loss function specific hyperparameters that were used are shown in Table \ref{tab:hyperP}. Here, we adopted the parameters used in the corresponding papers for Cifar-10 and Cifar-100. If not provided otherwise, we used the Cifar-10 parameters for Fashion-MNIST and the Cifar-100 parameters for WebVision.

Networks are initialized with zero biases and random Xavier-uniform entries \citep{Glorot.2010}  for the MLP networks, while the ResNet architectures use Kaiming Normal weights \citep{he2015delving}. We train the networks using five different initialization seeds. For robust results when comparing network performance, we report the mean of the accuracies over the five different network realizations, along with the corresponding errors of the mean.

\subsection{Choosing the right logit bias}
\begin{figure}
  \centering
  \includegraphics[width=0.9\linewidth]{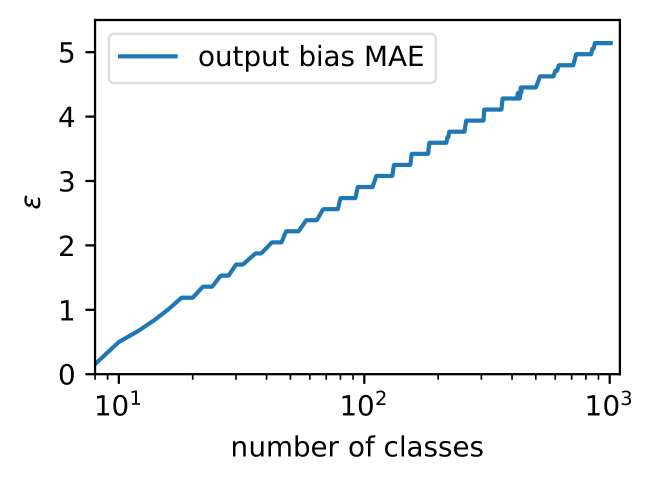}
  \caption{The logit bias value $\epsilon$ for the Mean Absolute Error as a function of the number of classes present in the dataset. }
  \label{fig:eps_error_c}
\end{figure}
In Eq.~\eqref{eq:constExpVal} we give a criterion that defines how $\epsilon$ may be chosen given the number of classes in the dataset, and an additional constant $C$ which is set to $0.15$ in the whole manuscript. 
In Figure \ref{fig:eps_error_c} we show the dependency of $\epsilon$ on the number of classes using this equation.

As illustrated in the manuscript for Cifar-100, slightly deviating from the calculated results can even improve performance. To further demonstrate the stability of a chosen $\epsilon$ we provide results on Cifar-100 with $20\%$ noise. The value chosen in the manuscript was $\epsilon=1.5$ and we find that the test-accuracy in dependence of the logit bias $[\epsilon=1.3: 67.9\%, \epsilon=1.4: 68.2\%, \epsilon=1.5: 68.3\%, \epsilon=1.6: 68.4\%, \epsilon=1.7: 68.2\% ]$ is stable around the value $\epsilon=1.5$.

On the WebVision dataset we found that having a rather large $\epsilon$ leads to low confidence predictions on the test set, as the network is used to the logit bias. To counteract this issue, we proposed to wean the network by also including epochs with a lower $\epsilon$ value.  This was done by taking 6 discrete $\epsilon$ values of $[5.2, 4.8, 4.3, 3.7, 3.0, 2.3]$ to iterate through during training. Training starts with $\epsilon=5.2$ and after 11 epochs, the next smaller $\epsilon$ is taken. After 66 epochs, the smallest $\epsilon$ is reached and the next epoch starts again with $\epsilon=5.2$. The reason for a decay that repeats after 66 epochs is the change in the learning rate after 66 and 132 epochs respectively. Hence, each learning rate gets an identical schedule. As this cycle completes after $198$ epochs we decided to only train for this amount of epochs in these cases.

In general we believe that a dynamically tuned $\epsilon$ could further improve performance even for datasets with fewer classes than WebVision. If the network gains confidence in a specific example in the later stages of training, the network bias either prevents full memorization of the example as in the case of 1000 classes or even allows an example with the wrong label to be learned even though the network was already fairly confident against it.
We have not included this in the present manuscript, since this approach could be considered to be fine-tuning of $\epsilon$, such that to ensure a fair comparison with other loss functions we would have to fine-tune the respective parameters as well, substantially increasing the compute time for numerical experiments. 

The problem of having to actively find the correct $\epsilon$ schedule could principally be mitigated by designing a different loss function that uses our insight into the early stages of learning. By choosing the derivative of the loss function to extend to the middle of the initial distribution independent of the class number we could most likely omit any schedule. However, this is a topic for future research.

\begin{figure}
  \centering
  \includegraphics[width=\linewidth]{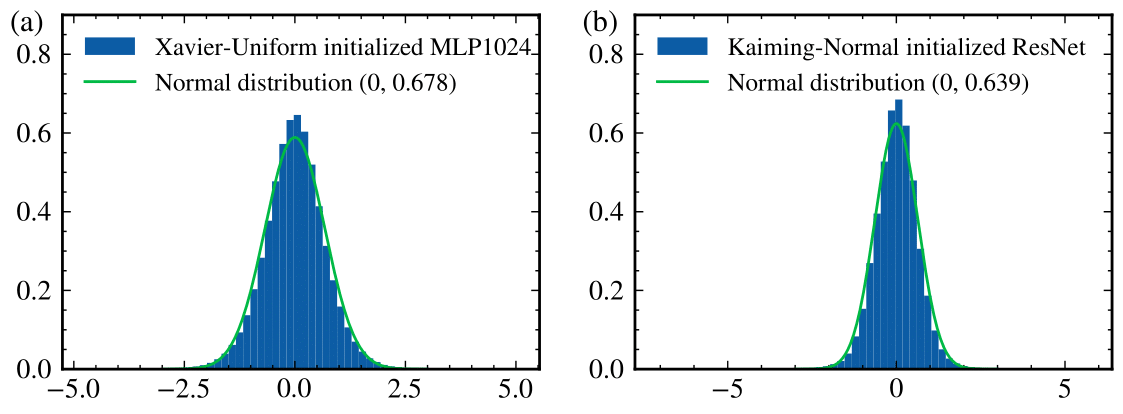}
  \caption{Logit distribution $p_z$ for (a) the fully connected network MLP1024 and (b) the ResNet-34 architecture. The histogram displays the empirical logits for different network initializations given the first images of the Cifar dataset. The green curve displays a normal distribution with a mean of zero. The variance of the normal distribution is set to the variance of the logit distribution.   }
  \label{fig:output_networks}
\end{figure}

\subsection{Empirical output of a freshly initialized neural network}
In the main text, we derive that the logit distribution of a freshly initialized neural network $p_z$ follows a normal distribution with a mean of zero if the weights are initialized with a mean of zero. The standard deviation depends on the distribution which was used to initialize the weights. This is verified in Figure \ref{fig:output_networks} where panel (a) shows the empirical logits of $500$ Xavier-uniform initialized MLP1024 networks when given the first $150$ Cifar-10 images. Panel (b) shows the logits of $20$ different ResNet-34 networks initialized with Kaiming-Normal distributed numbers given the first $300$ images of the Cifar-100 dataset. 

\begin{figure}

\begin{minipage}[c]{0.45\linewidth}
\includegraphics[width=\linewidth]{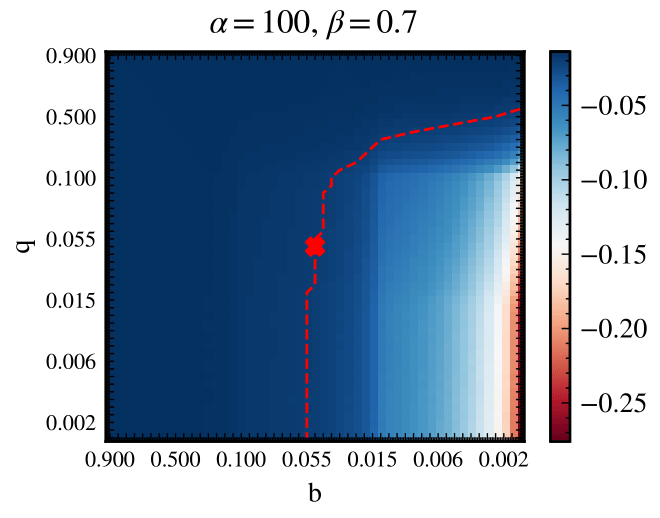}
\end{minipage}
\hfill
\begin{minipage}[c]{0.45\linewidth}
\includegraphics[width=\linewidth]{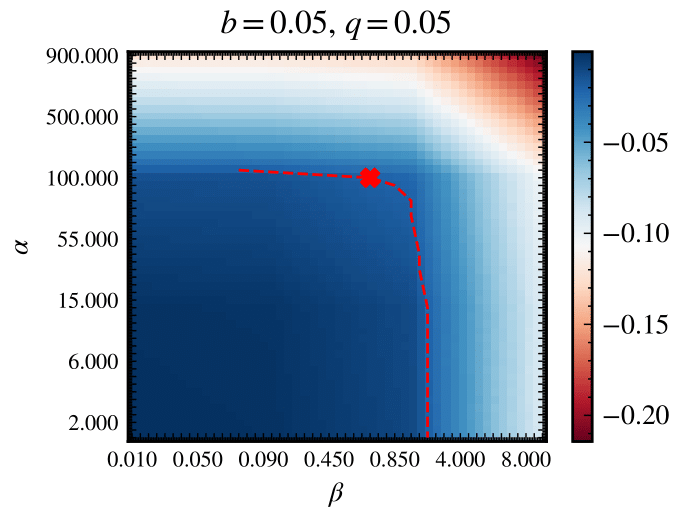}
\end{minipage}%
\caption{Optimization of the four parameters of the NCE-AGCE loss. We display the error $\langle \delta_k \rangle (0)$ as a function of $q$ and $b$ (left pannel) and as a function of $\alpha$ and $\beta$ (right pannel). The dashed red line displays possible solutions to the equation $\langle \delta_k \rangle (0)=  -0.021$ to keep the learning comparable to learning on the Cifar-100 dataset. The red cross shows the chosen solution, where the decision is based on a learning curve plot.  \label{fig:optiNCE-AGCE} }
\end{figure}

\begin{table*}[t!]
  \centering 
  \begin{tabular}{l l l l l l l l }
    \toprule 
    Dataset, Network   &Loss      & \multicolumn{1}{c}{Noise: 0 \%\,\,\,}    & \multicolumn{1}{c}{10 \%} & \multicolumn{1}{c}{20 \%}  & \multicolumn{1}{c}{40 \%} & \multicolumn{1}{c}{60 \%}  \\
    \midrule  
         & CE                  & \textbf{55.70} $\pm$ 0.13          & 50.56 $\pm$ 0.10          & 46.20 $\pm$ 0.31          & 36.90 $\pm$ 0.17          & 27.51 $\pm$ 0.35         \\
         &  MAE                 & 49.86 $\pm$ 1.13          & 49.08 $\pm$ 0.91          & 48.20 $\pm$ 1.04          & 45.31 $\pm$ 0.93          & 40.27 $\pm$ 1.06         \\
Cifar-10   &  MAE* $\epsilon=0.5$  & 55.09 $\pm$ 0.14          & \textbf{53.83} $\pm$ 0.15          & \textbf{53.18} $\pm$ 0.20 & 49.30 $\pm$ 0.08          & 42.29 $\pm$ 0.13         \\
         &  genCE               & 55.21 $\pm$ 0.13          & \textbf{54.07} $\pm$ 0.22 & 52.72 $\pm$ 0.14          & 47.94 $\pm$ 0.13          & 36.70 $\pm$ 0.26         \\
MLP1024  &  symCE               & 54.13 $\pm$ 0.14          & 49.03 $\pm$ 0.17          & 44.30 $\pm$ 0.13          & 34.50 $\pm$ 0.19          & 25.23 $\pm$ 0.15         \\
         &  NF-MAE             & 53.98 $\pm$ 0.13          & 53.20 $\pm$ 0.15          & 52.00 $\pm$ 0.08          & \textbf{49.90} $\pm$ 0.18 & \textbf{45.43} $\pm$ 0.17\\
         &  NCE-MAE             & 53.77 $\pm$ 0.10          & 53.10 $\pm$ 0.05          & 52.18 $\pm$ 0.12          & \textbf{49.62} $\pm$ 0.24          & 44.88 $\pm$ 0.21         \\
         &  biTemp              & \textbf{56.02} $\pm$ 0.24 & 51.05 $\pm$ 0.11          & 46.58 $\pm$ 0.18          & 37.35 $\pm$ 0.21          & 27.28 $\pm$ 0.22         \\
     \midrule 
              &  CE                  & \textbf{90.20} $\pm$ 0.03 & 86.05 $\pm$ 0.04          & 79.54 $\pm$ 0.13          & 63.04 $\pm$ 0.34          & 45.67 $\pm$ 0.39         \\
              &  MAE                 & 84.83 $\pm$ 2.59          & 81.25 $\pm$ 1.44          & 82.00 $\pm$ 2.02          & 83.70 $\pm$ 1.58          & 80.17 $\pm$ 2.60         \\
    Fashion-MNIST  &  MAE* $\epsilon=0.5$  & 89.55 $\pm$ 0.03          & 89.48 $\pm$ 0.05          & \textbf{89.09} $\pm$ 0.07          & \textbf{88.03} $\pm$ 0.07 & 85.63 $\pm$ 0.12         \\
              &  genCE               & 89.76 $\pm$ 0.07          & \textbf{89.49} $\pm$ 0.05 & \textbf{89.09} $\pm$ 0.10 & 87.70 $\pm$ 0.12          & 80.86 $\pm$ 0.19         \\
MLP1024       &  symCE               & 89.98 $\pm$ 0.07          & 88.06 $\pm$ 0.04          & 84.05 $\pm$ 0.13          & 69.95 $\pm$ 0.37          & 51.34 $\pm$ 0.20         \\
              &  NF-MAE             & 89.13 $\pm$ 0.03          & 88.99 $\pm$ 0.04          & 88.62 $\pm$ 0.06          & \textbf{87.86} $\pm$ 0.10          & \textbf{86.26} $\pm$ 0.08\\
              &  NCE-MAE             & 89.14 $\pm$ 0.03          & 88.95 $\pm$ 0.06          & 88.59 $\pm$ 0.05          & 87.79 $\pm$ 0.08          & \textbf{86.24} $\pm$ 0.06         \\
              &  biTemp              & \textbf{90.19} $\pm$ 0.08          & 86.46 $\pm$ 0.11          & 80.33 $\pm$ 0.15          & 64.11 $\pm$ 0.37          & 45.77 $\pm$ 0.29         \\
    \bottomrule 
  \end{tabular}
    \caption{Final test accuracies for training MLP1024 networks on the ten-class datasets Cifar-10 and Fashion-MNIST with different loss functions and various amounts of label noise. For the optimized learning rates see Tab.~\ref{tab:lr_optimization}.  }
  \label{pred_table_5}
\end{table*} 

\subsection{Estimating hyperparameters of other loss functions}
Other loss functions often use hyperparameters to adjust for different learning scenarios. The various parameters used in the manuscript are summarized in appendix Tab.~\ref{tab:hyperP} and show that their behaviour can be highly non trivial. To reduce hyperparameters we propose to instead use the overlap shown in Fig.~\ref{fig:learning_curves} to adjust these parameters. 
Using parameters which are known to work with a specific number of classes, one computes the value of the backpropagation error $\langle\delta_k\rangle (z_k)$ at the position $z_k = 0$ for this number of classes. When changing the number of classes, the hyperparameters are adjusted until $\langle\delta_k\rangle(0)$ coincides with the value of the baseline for which it is known that learning works well. 

As an example, we consider genCE where we use that for $C=100$ classes the value $q_{100}=0.7$ has been established and compute the value of the backpropagation error $\langle\delta_k\rangle (0)$ to be around $-0.279$. 
For $C=1000$ classes, we find that $q_{1000}=0.48$ exactly mimics this behavior. 

In the case of multiple parameters there are potentially many solutions to the equation. Finding a \emph{good}
solution may require looking at the curves of the backpropagation error, as shown in
Fig.~\ref{fig:learning_curves},
because there may be trivial but misleading solutions. For example, in the case of a bounded loss function with a scaling parameter, simply scaling the loss by a factor of $100$ might solve the equation. However, this will most likely lead to exploding gradients in other phases of learning. 
For the NCE-AGCE loss we found the suitable parameters
$\alpha=100$, $\beta=0.7$, $q=0.05$ and $b=0.05$, which fulfill the relation that $\langle\delta_k\rangle(0) = -0.021$ as for the case of $100$ classes. The parameter space with potential solutions is displayed in Fig.~\ref{fig:optiNCE-AGCE} where we display $\langle \delta_k \rangle (0)$ for various parameter combinations. Potential solutions are marked by the dashed red line, while the chosen solution is indicated by a cross. The decision for this solution is based on the resulting learning curve.

\subsection{Empirical results on a different architecture}
To train MLP1024 networks on the Cifar-10 and Fashion-MNIST datasets, we perform a grid search over nine values of initial learning rates in the absence of label noise. The value with the best average accuracy out of five runs for each learning rate is then used for training in the presence of varying degrees of label noise. The results for the learning rate optimization are shown in Table \ref{tab:lr_optimization}, where values in bold font mark the learning rate used for the simulations with noise.

The final results when training MLP1024 Networks on Cifar-10 and Fashion-MNIST are summarized in Tab.~\ref{pred_table_5}, where we find that MAE* performed best under all loss functions.

\subsection{Backpropagation error of bounded losses}
For completeness, we provide the backpropagation errors $\delta_n$ for the active-passive losses and AGCE-NCE in this section. As in the main text, $k$ denotes the index of the non-zero entry in the corresponding one-hot encoded label, i.e.\@ $k={\rm argmax}(\bm{y})$.

The active passive losses are defined as
\begin{align}
   {\rm NF-MAE} &= \alpha \; \frac{ \log \left( (1- a_k)^{0.5} a_k  \right) }{ \sum_i \log \left(  (1- a_i)^{0.5} a_i \right) } + \beta \; {\rm MAE}(\bm{a},\bm{y}) \ , \\
   {\rm NCE-MAE} &= \alpha \; \frac{ \log(a_k) }{ \sum_i \log(  a_i ) } + \beta \; {\rm MAE}(\bm{a},\bm{y}) \ .
\end{align}
We are interested in the error $\delta_n = \partial_{z_n}\mathcal{L}$ where $a_i = \exp(z_n) / \sum_i \exp z_i$. The derivative of the mean absolute error (MAE) is given in the main text 
as $2a_k(a_k-\delta[n-k])$.
We further find that 
\begin{align}
\begin{split}
    &\frac{\partial}{\partial z_n} \frac{ \log \left( (1- a_k)^{0.5} a_k  \right) }{ \sum_i \log \left(  (1- a_i)^{0.5} a_i \right) } =  \\
    &\frac{  (\delta[n-k]-a_n)  \left( 1- \frac{a_k}{2-2a_k} \right) \left( \sum_i \log( (1-a_k)^{0.5} a_k ) \right)}{\left( \sum_i \log((1-a_i)^{0.5} a_i) \right)^2} - \\
    &\frac{ \left( \sum_i (\delta[n-i] -a_n) (1- \frac{a_i }{2-2 a_i} \right) \log( (1-a_k)^{0.5} a_k) }
{\left( \sum_i \log((1-a_i)^{0.5} a_i) \right)^2} \ , \\
     &\frac{\partial}{\partial z_n} \frac{ \log(a_k) }{ \sum_i \log(  a_i ) } = \frac{(\delta[n-k] -a_n) \sum_i \log(a_i) - \log(a_k) ( -c a_n + 1)} { \left( \sum_i \log(a_i) \right)^2  } \ .
    \end{split}
\end{align}

For the AGCE-NCE loss we simply combine the already found derivatives:
\begin{align}
\begin{split}
    &\frac{\partial}{\partial z_n} \frac{ \log(a_k) }{ \sum_i \log(  a_i ) } + [(b+1)^q-(b +a_k)^q]/q = \\
    &\frac{(\delta[n-k] -a_n) \sum_i \log(a_i) - \log(a_k) ( -c a_n + 1)} { \left( \sum_i \log(a_i) \right)^2  } + \\  
    & a_k (a_n -\delta[n-k] )(b+a_n)^{q-1}
    \end{split}
\end{align}
 A visualization of these formulas is provided in the main manuscript.

\begin{table*} 
  \centering 
  \begin{tabular}{l c c c c c }
    \toprule   
    Cifar-10 \\
    \midrule
    learning rate & 0.0005 & 0.0008 & 0.001 & 0.003 & 0.005   \\
    \midrule  
CE  & 54.42 $\pm$ 0.06 & 55.01 $\pm$ 0.10 & \textbf{55.70} $\pm$ 0.13 & 55.41 $\pm$ 0.15 & 55.03 $\pm$ 0.09 \\
MAE   & 49.44 $\pm$ 1.11 & \textbf{49.86} $\pm$ 1.13 & 49.18 $\pm$ 1.82 & 43.03 $\pm$ 1.26 & 22.80 $\pm$ 2.81  \\
MAE* $\epsilon=0.5$  & 52.98 $\pm$ 0.13 & 53.84 $\pm$ 0.05 & 54.20 $\pm$ 0.09 & \textbf{55.09} $\pm$ 0.14 & 42.67 $\pm$ 3.46  \\
genCE  & 52.54 $\pm$ 0.16 & 53.54 $\pm$ 0.16 & 53.91 $\pm$ 0.07 & 54.99 $\pm$ 0.18 & \textbf{55.21} $\pm$ 0.13 \\
NF-MAE  & 53.67 $\pm$ 0.15 & 53.88 $\pm$ 0.10 & \textbf{53.98} $\pm$ 0.13 & 16.48 $\pm$ 2.93 & 14.92 $\pm$ 1.88  \\
NCE-MAE  & 53.82 $\pm$ 0.04 & \textbf{53.77} $\pm$ 0.10 & 53.49 $\pm$ 0.21 & 16.61 $\pm$ 0.81 & 14.12 $\pm$ 1.10 \\
biTemp  & 53.17 $\pm$ 0.15 & 54.23 $\pm$ 0.07 & 54.69 $\pm$ 0.18 & 55.89 $\pm$ 0.22 & \textbf{56.02} $\pm$ 0.24 \\
    \midrule 
    learning rate & 0.008 & 0.01 & 0.03 & 0.05  \\
    \midrule 
CE   & 54.65 $\pm$ 0.10 & 54.44 $\pm$ 0.13 & 10.00 $\pm$ 0.00 & 10.00 $\pm$ 0.00 \\
MAE   & 11.37 $\pm$ 0.97 & 10.92 $\pm$ 0.92 & 10.00 $\pm$ 0.00 & 10.00 $\pm$ 0.00 \\
MAE* $\epsilon=0.5$   & 10.35 $\pm$ 0.35 & 12.20 $\pm$ 0.97 & 10.00 $\pm$ 0.00 & 10.00 $\pm$ 0.00 \\
genCE   & 55.02 $\pm$ 0.26 & 54.57 $\pm$ 0.07 & 10.00 $\pm$ 0.00 & 10.00 $\pm$ 0.00 \\
NF-MAE   & 10.00 $\pm$ 0.00 & 10.83 $\pm$ 0.83 & 10.46 $\pm$ 0.46 & 10.00 $\pm$ 0.00 \\
NCE-MAE  & 13.09 $\pm$ 1.04 & 11.43 $\pm$ 1.43 & 10.00 $\pm$ 0.00 & 10.00 $\pm$ 0.00 \\
biTemp   & 55.21 $\pm$ 0.17 & 54.97 $\pm$ 0.18 & 50.76 $\pm$ 0.25 & 28.66 $\pm$ 8.11 \\
\toprule
Fashion-MNIST \\
\midrule
 learning rate & 0.0005 & 0.0008 & 0.001 & 0.003 & 0.005   \\
 \midrule
CE & 89.39 $\pm$ 0.06 & 89.63 $\pm$ 0.08 & 89.64 $\pm$ 0.08 & 90.11 $\pm$ 0.05 & \textbf{90.20} $\pm$ 0.03 \\
MAE & 81.24 $\pm$ 2.24 & \textbf{84.83} $\pm$ 2.32 & 82.53 $\pm$ 2.60 & 82.11 $\pm$ 2.21 & 84.34 $\pm$ 2.00 \\
MAE* $\epsilon=0.5$ & 88.72 $\pm$ 0.05 & 88.52 $\pm$ 0.95 & 88.52 $\pm$ 0.95 & \textbf{89.55} $\pm$ 0.03 & 88.52 $\pm$ 0.95 \\
genCE & 88.54 $\pm$ 0.05 & 88.82 $\pm$ 0.03 & 88.93 $\pm$ 0.04 & 89.57 $\pm$ 0.05 & 89.60 $\pm$ 0.08 \\
NF-MAE & 88.73 $\pm$ 0.05 & 89.06 $\pm$ 0.04 & \textbf{89.13} $\pm$ 0.03 & 88.30 $\pm$ 0.03 & 30.28 $\pm$ 4.61 \\
NCE-MAE & 88.82 $\pm$ 0.03 & 89.05 $\pm$ 0.03 & \textbf{89.14} $\pm$ 0.03 & 88.31 $\pm$ 0.05 & 43.06 $\pm$ 8.93 \\
biTemp & 89.08 $\pm$ 0.03 & 89.33 $\pm$ 0.10 & 89.50 $\pm$ 0.06 & 89.97 $\pm$ 0.06 & 90.13 $\pm$ 0.06 \\
\midrule
learning rate & 0.008 & 0.01 & 0.03 & 0.05  \\
\midrule
CE & 89.97 $\pm$ 0.05 & 90.08 $\pm$ 0.06 & 89.19 $\pm$ 0.10 & 10.00 $\pm$ 0.00\\
MAE & 79.61 $\pm$ 2.65 & 72.00 $\pm$ 1.88 & 10.03 $\pm$ 0.03 & 12.00 $\pm$ 1.79\\
MAE* $\epsilon=0.5$ & 89.43 $\pm$ 0.06 & 87.55 $\pm$ 0.52 & 10.00 $\pm$ 0.00 & 10.01 $\pm$ 0.01\\
genCE & 89.75 $\pm$ 0.06 & \textbf{89.76} $\pm$ 0.06 & 17.00 $\pm$ 1.60 & 11.06 $\pm$ 0.95\\
NF-MAE & 20.34 $\pm$ 7.53 & 17.47 $\pm$ 1.70 & 10.00 $\pm$ 0.00 & 12.36 $\pm$ 1.57\\
NCE-MAE & 10.00 $\pm$ 0.00 & 13.94 $\pm$ 2.16 & 15.95 $\pm$ 2.17 & 11.87 $\pm$ 1.66\\
biTemp & 90.16 $\pm$ 0.07 & \textbf{90.19} $\pm$ 0.07 & 89.93 $\pm$ 0.06 & 89.44 $\pm$ 0.09\\
    \bottomrule 
  \end{tabular}
    \caption{ Learning rate optimization for various loss functions on the clean dataset for MLP1024 networks. The test accuracies in boldface indicate the learning rate for which results are presented in the main manuscript.  MAE* has $\epsilon=0.5$ }
\label{tab:lr_optimization}
\end{table*}

\end{document}